\documentclass[letterpaper]{llncs}
\usepackage{times}
\usepackage{helvet}
\usepackage{courier}
\usepackage{epic}
\usepackage{graphicx}
\usepackage{latexsym}
\usepackage{verbatim}
\usepackage{xspace}
\begin{document}

\newtheorem{mydefinition}{Definition}
\newtheorem{mytheorem}{Theorem}
\newtheorem{mycor}{Corollary}
\newenvironment{myexample}{{\bf Example:} \it}{\rm}
\newtheorem{mytheorem1}{Theorem}
\newcommand{\myproof}{\noindent {\bf Proof:\ \ }}
\newcommand{\myqed}{\mbox{$\Box$}}

\newcommand{\mymod}{\mbox{\rm mod}}
\newcommand{\range}{\mbox{\sc Range}}
\newcommand{\roots}{\mbox{\sc Roots}}
\newcommand{\myiff}{\mbox{\rm iff}}
\newcommand{\alldifferent}{\mbox{\sc AllDifferent}}
\newcommand{\alldiff}{\mbox{\sc AllDifferent}}
\newcommand{\interdistance}{\mbox{\sc InterDistance}}
\newcommand{\permutation}{\mbox{\sc Permutation}}
\newcommand{\disjoint}{\mbox{\sc Disjoint}}
\newcommand{\cardpath}{\mbox{\sc CardPath}}
\newcommand{\CARDPATH}{\mbox{\sc CardPath}}
\newcommand{\knapsack}{\mbox{\sc Knapsack}}
\newcommand{\common}{\mbox{\sc Common}}
\newcommand{\uses}{\mbox{\sc Uses}}
\newcommand{\lex}{\mbox{\sc Lex}}
\newcommand{\usedby}{\mbox{\sc UsedBy}}
\newcommand{\nvalue}{\mbox{\sc NValue}}
\newcommand{\slide}{\mbox{\sc Slide}}
\newcommand{\SLIDE}{\mbox{\sc Slide}}
\newcommand{\circularslide}{\mbox{\sc Slide}_{\rm O}}
\newcommand{\among}{\mbox{\sc Among}}
\newcommand{\mysum}{\mbox{\sc Sum}}
\newcommand{\amongseq}{\mbox{\sc AmongSeq}}
\newcommand{\atmost}{\mbox{\sc AtMost}}
\newcommand{\atleast}{\mbox{\sc AtLeast}}
\newcommand{\element}{\mbox{\sc Element}}
\newcommand{\gcc}{\mbox{\sc Gcc}}
\newcommand{\egcc}{\mbox{\sc EGcc}}
\newcommand{\gsc}{\mbox{\sc Gsc}}
\newcommand{\contiguity}{\mbox{\sc Contiguity}}
\newcommand{\PRECEDENCE}{\mbox{\sc Precedence}}
\newcommand{\GENPRECEDENCE}{\mbox{\sc GenPrecedence}}
\newcommand{\assignnvalues}{\mbox{\sc Assign\&NValues}}
\newcommand{\linksettobooleans}{\mbox{\sc LinkSet2Booleans}}
\newcommand{\domain}{\mbox{\sc Domain}}
\newcommand{\symalldiff}{\mbox{\sc SymAllDiff}}
\newcommand{\valsymbreak}{\mbox{\sc LexLeader}}
\newcommand{\valsym}{\mbox{\sc LexLeader}}
\newcommand{\LexLeader}{\mbox{\sc LexLeader}}

\newcommand{\slidingsum}{\mbox{\sc SlidingSum}}
\newcommand{\MaxIndex}{\mbox{\sc MaxIndex}}
\newcommand{\REGULAR}{\mbox{\sc Regular}}
\newcommand{\regular}{\mbox{\sc Regular}}
\newcommand{\STRETCH}{\mbox{\sc Stretch}}
\newcommand{\SLIDEOR}{\mbox{\sc SlideOr}}
\newcommand{\NAE}{\mbox{\sc NotAllEqual}}
\newcommand{\mymax}{\mbox{\rm max}}
\newcommand{\todo}[1]{{\tt (... #1 ...)}}

\newcommand{\DC}{\ensuremath{DC}\xspace}
\newcommand{\Xbf}{\mbox{{\bf X}}\xspace}
\newcommand{\LEXCHAIN}{\mbox{\sc LexChain}}
\newcommand{\DLex}{\mbox{\sc DoubleLex}\xspace}
\newcommand{\snakelex}{\mbox{\sc SnakeLex}\xspace}
\newcommand{\DLexColSum}{\mbox{\sc DoubleLexColSum}\xspace}

\title{Parameterized Complexity Results\\ in Symmetry Breaking\thanks{Supported by
the Australian 
Government's  Department of Broadband, Communications and the Digital Economy
and the
ARC. I wish to thank my co-authors in this area for their many
contributions: Christian Bessiere,  Brahim Hnich,
Emmanuel Hebrard, George Katsirelos, Zeynep Kiziltan, Yat-Chiu Law, 
               Jimmy Lee,  Nina Narodytska
               Claude-Guy Quimper and  Justin Yip.}}

\author{Toby Walsh}
\institute{NICTA and University of New South Wales, email:
toby.walsh@nicta.com.au}

\maketitle
\begin{abstract}
Symmetry is a common feature of many combinatorial
problems. Unfortunately eliminating all symmetry
from a problem is often computationally intractable.
This paper argues that recent parameterized complexity results
provide insight into that intractability and help
identify special cases in which symmetry can be dealt
with more tractably. 
\end{abstract}

\section{Introduction}

Symmetry occurs in many constraint satisfaction problems. 
For example, in scheduling a round robin sports tournament,
we may be able to interchange all the matches taking
place in two stadia. Similarly, we may be able to
interchange two teams throughout the tournament. 
As a second example, 
when colouring a graph (or equivalently when
timetabling exams), the colours are interchangeable. 
We can swap red with blue throughout. If we 
have a proper colouring, any permutation
of the colours is itself a proper colouring.
Problems may have many symmetries at once. 
In fact, the symmetries of a problem form a group. Their
action is to map solutions (a schedule, 
a proper colouring, etc.) onto solutions. 

Symmetry is problematic when solving 
constraint satisfaction problems as we may waste much time visiting
symmetric solutions. In addition,
we may visit many (failing) search states that
are symmetric to those that we have already visited. 
One simple but effective mechanism to deal with symmetry is to
add constraints which eliminate symmetric solutions \cite{puget:Sym}.
Unfortunately eliminating all symmetry is
NP-hard in general \cite{clgrkr96}. However,
recent results in parameterized complexity
give us a good understanding
of the source of that complexity. 
In this survey paper, I summarize results
in this area. For more background, see \cite{wpricai2010,kwecai2010,hwaaai2010,waaai2008,wai06}. 

\section{An example}

To illustrate the ideas, we consider a simple
problem from musical composition. 
The all interval series problem (prob007 in CSPLib.org \cite{csplib})
asks for a permutation 
of the numbers 0 to $n-1$
so that neighbouring differences
form a permutation of 1 to $n-1$. 
For $n=12$, the problem corresponds to 
arranging the half-notes of a scale
so that all musical intervals (minor second to
major seventh) are covered. 
This is a simple example of a graceful graph
problem in which the graph is a path.
We can model
this as a constraint satisfaction problem
in $n$ variables with
$X_i=j$ iff the $i$th number in the series is $j$. 
One solution for $n=11$ is:
\begin{eqnarray}
X_1, X_2, \ldots, X_{11} & = & 
3, 7, 4, 6, 5, 0, 10, 1, 9, 2, 8 
\end{eqnarray}
The differences form the series: $ 4, 3, 2, 1, 5, 10, 9, 8, 7, 6 $.

The all interval series problem has a number of different symmetries.
First, we can reverse any solution and generate a new
(but symmetric) solution:
\begin{eqnarray}
X_1, X_2,   \ldots, X_{11} & = & 
 8, 2, 9, 1, 10, 0, 5, 6, 4, 7, 3
\end{eqnarray}
Second, the all interval series problem has a value symmetry 
as we can invert values.
If we subtract all values in (1) from $10$, we
generate a second (but symmetric) solution:
\begin{eqnarray}
X_1,  X_2,  \ldots, X_{11} & = & 
 7, 3, 6, 4, 5, 10, 0, 9, 1, 8, 2
\end{eqnarray}
Third, we can do both and generate a third (but symmetric)
solution:
\begin{eqnarray}
X_1,  X_2,  \ldots, X_{11} & = & 
 2, 8, 1, 9, 0, 10, 5, 4, 6, 3, 7
\end{eqnarray}

To eliminate such symmetric solutions from the
search space, we can post additional constraints
which eliminate all but one solution in each
symmetry class. 
To eliminate the reversal of a solution,
we can simply post the constraint:
\begin{eqnarray}
& X_1 < X_{11}&
\end{eqnarray}
This eliminates solution (2) as it is a reversal of (1). 
To eliminate the value symmetry which subtracts all values
from $10$, we can post:
\begin{eqnarray}
X_1 \leq 5, & & X_1=5 \Rightarrow X_2 < 5
 \end{eqnarray}
This eliminates solutions (2) and (3). 
Finally, eliminating the third symmetry
where we both reverse the solution and subtract it
from $10$ is more difficult. We can, for instance, post:
\begin{eqnarray}
[ X_1, \ldots , X_{11}]  & \leq_{\rm lex} & 
[10-X_{11}, \ldots, 10-X_{1}]
\end{eqnarray}
Note that of the four symmetric solutions given
earlier, only (4)
with $X_1=2$, $X_2=8$ and $X_{11}=7$
satisfies all three sets of symmetry breaking constraints: (5), (6) and (7).
The other three solutions are eliminated.

\section{Formal background}

We will need some formal notation to present some of the
more technical results. 
A \emph{constraint satisfaction problem} (CSP) consists of a set of
variables, each with a finite domain of values, and a set of
constraints \cite{fvwhandbook}. Each \emph{constraint} is 
specified by the allowed combinations of values for some
subset of variables. For example, $X \neq Y$ is 
a binary constraint which ensures $X$ and $Y$ do not
take the same values. 
A \emph{global constraint} is one in which the number of variables
is not fixed. 
For instance,
the global constraint $\nvalue([X_1,\ldots, X_n],N)$ ensures that $n$ variables, $X_1$ to $X_n$,
take $N$ different values \cite{pachet1}. 
That is, $N = | \{ X_i \ | 1 \leq i \leq n\}|$. 

Constraint solvers typically use
backtracking search to explore the space
of partial assignments. After each assignment,
constraint propagation algorithms prune the search
space by enforcing local consistency properties like domain or bound
consistency. A constraint is \emph{domain
consistent} (\emph{DC})
iff when a variable is assigned any of the values in its domain, there
exist compatible values in the domains of all the other variables of
the constraint. Such values are called
a \emph{support}. A CSP is domain consistent iff every
constraint is domain consistent.

Recently, Bessiere {\it et al.} have
shown that a number of common global constraints
are intractable to propagate 
\cite{bhhwaaai2004,bhhwcp04}.
For instance, enforcing
domain consistency on the \nvalue\ constraint
is NP-hard
\cite{bhhkwconstraint2006,bhhkwcpaior2005}. 
Parameterized complexity
can provide a more fine-grained
view of such results,
identifying
more precisely what makes a global constraint
(in)tractable. 
We will say that a problem is \emph{fixed-parameter tractable}
(\emph{FPT}) if it can be solved in $O(f(k) n^c)$ time where
$f$ is \emph{any} computable function, $k$ is
some parameter, $c$ is a constant, and $n$ is
the size of the input.
For example, vertex cover (``Given a graph with $n$ vertices, is there
a subset of vertices of size $k$ or less that
cover each edge in the graph'') is NP-hard in general,
but fixed-parameter tractable with respect to $k$ since it
can be solved in $O({1.31951}^k k^2 + kn)$ time \cite{DowFelSte99}.
Hence, provided $k$ is small, vertex cover
can be solved effectively.

\section{Symmetry breaking}

As we have argued, symmetry is a common feature of many
real-world problems that dramatically increases the size
of the search space if it is not
factored out. 
Symmetry can be defined as a bijection
on assignments that preserves solutions.
The set of symmetries form a group under
the action of composition. 
We focus on two special types of symmetry.
A \emph{value symmetry} is a bijective mapping, $\sigma$ of the
values such that if $X_1=d_1, \ldots, X_n=d_n$ is a solution
then $X_1=\sigma(d_1), \ldots, X_n=\sigma(d_n)$ is also. 
For example, in our all interval series example,
there is a value symmetry $\sigma$ that maps
the value $i$ onto $n-i$. 
A \emph{variable symmetry}, on the other hand,
is a bijective mapping, $\theta$ of the
indices of variables such that if $X_1=d_1, \ldots, X_n=d_n$ is a solution
then $X_{\theta(1)}=d_1, \ldots, X_{\theta(n)}=d_n$ is also. 
For example, in our all interval series example,
there is a variable symmetry $\theta$ that maps
the index $i$ onto $n+1-i$. This swaps
the variable $X_i$ with $X_{n+1-i}$. 

A simple and effective
mechanism to deal with symmetry is to
add constraints to eliminate symmetric solutions
\cite{puget:Sym,clgrkr96,wcp06,knwercim08,llwycp07,hkwaimath04}.
The basic idea is very simple. 
We pick an ordering on the variables, and
then post symmetry breaking constraints to
ensure that the final solution is lexicographically
less than any symmetric re-ordering of the variables. 
That is, we select the ``lex leader'' assignment. 
For example, to break the variable symmetry $\theta$,
we post the constraint:
\begin{eqnarray*}
[ X_1, \ldots , X_n]  & \leq_{\rm lex} & 
[X_{\theta(1)}, \ldots, X_{\theta(n)}]
\end{eqnarray*}
Efficient inference methods exist for propagating such constraints 
\cite{fhkmwcp2002,fhkmwaij06}. 
The symmetry breaking constraints in 
our all interval series example can all
be derived from such lex leader constraints.

In theory, the lex leader method solves the problem of symmetries,
eliminating all symmetric solutions and pruning
many symmetric states. Unfortunately, the set of 
symmetries might be exponentially large (for example,
in a  graph $k$-colouring, there are $k!$ symmetries). There 
may therefore be too many symmetry breaking 
constraints to post. In addition, decomposing
symmetry breaking into many lex leader constraints
typically hinders propagation. 
We focus on three special but
commonly occurring cases where 
symmetry breaking is more tractable and
propagation can be more powerful:
value symmetry, interchangeable
values, and row and column
symmetry. In each case, we identify
islands of tractability but show that
the quick-sands of intractability
remain close to hand. 

\section{Value symmetry}

Value symmetries are a commonly occurring
symmetry that are more tractable to break \cite{waaai2008}.
For instance, Puget has proved that a linear
number of symmetry breaking constraints will eliminate any
number of value symmetries in polynomial time \cite{pcp05}.
Given a set of value symmetries $\Sigma$,
we can eliminate all value symmetry
by posting the global constraint
$\valsymbreak(\Sigma,[X_1,\ldots,X_n])$
\cite{wcp06}.
This is a conjunction of lex leader constraints,
ensuring that, for each $\sigma \in \Sigma$:
$$\langle X_1, \ldots, X_n \rangle \leq_{\rm lex}
\langle \sigma(X_1), \ldots, \sigma(X_n) \rangle $$

Unfortunately, 
enforcing domain consistency on this
global constraint is NP-hard.
However, this complexity depends on the number
of symmetries.
Breaking all value symmetry is fixed-parameter tractable
in the number of symmetries.

\begin{mytheorem}
Enforcing domain consistency
on $\valsymbreak(\Sigma,[X_1,\ldots,X_n])$
is NP-hard in general but
fixed-parameter tractable
in $k = | \Sigma|$.
%
\end{mytheorem}
\myproof
NP-hardness is proved by Theorem 1 in \cite{wcp07}, 
and fixed-parameter tractability by Theorem 7 in \cite{bhhkqwaaai2008}.
\myqed

One situation where we may have only a small
number of symmetries is when we focus
on just the \emph{generators} of the symmetry
group \cite{clgrkr96,armsdac2002}. This is attractive
as the size of the generator set is logarithmic
in the size of the group, 
many algorithms in computational group
theory work on generators, and
breaking just the generator symmetries has been
shown to be effective on many benchmarks
\cite{armsdac2002}.
In general, breaking just the generators
may leave some symmetry. However, on
certain symmetry groups (like that for
interchangeable values considered in the
next section), all symmetry is eliminated (Theorem 3 in \cite{wcp07}).

\section{Interchangeable values}

By exploiting special properties of the value symmetry
group, we can identify even more tractable cases.
A common type of value symmetry with such properties
is that due to interchangeable values.
We can break all such symmetry
using the idea of value {\em precedence} \cite{llcp2004}.
In particular, we can post 
the global symmetry
breaking constraint $\PRECEDENCE([X_1,\ldots ,X_n])$.
This ensures
that for all $j<k$:
$$\min \{ i \ | \ X_i=j \vee i=n+1\} < 
 \min \{ i \ | \ X_i=k \vee i=n+2\}$$
That is, 
the first time we use $j$ is before
the first time we use $k$ for all $j<k$. 
For example, consider the assignment:
\begin{eqnarray}
X_1, X_2, X_3, \ldots, X_n & = & 1, 1, 1, 2, 1, 3, \ldots, 2
\end{eqnarray}
This satisfies value precedence as 1 first occurs
before 2, 2 first occurs before 3, etc. 
Now consider the symmetric assignment in which we swap 
$2$ with $3$:
\begin{eqnarray}
X_1, X_2, X_3, \ldots, X_n & = & 1, 1, 1, 3, 1, 2, \ldots, 3
\end{eqnarray}
This does not satisfy value precedence as 
$3$ first occurs before $2$. 
A \PRECEDENCE\ constraint 
eliminates all symmetry due to interchangeable values. 
In \cite{wecai2006}, we give a linear time propagator
for enforcing domain consistency on the \PRECEDENCE\ constraint.
In \cite{wcp07}, 
we argue that $\PRECEDENCE$ 
can be derived from the lex leader method (but offers
more propagation by being a global constraint). 

Another way to ensure value precedence
is to map onto dual variables, $Z_j$ which record the
first index using each value $j$
\cite{pcp05}. 
This transforms value symmetry into variable
symmetry on the $Z_j$. We can then eliminate
this variable symmetry with some ordering constraints:
\begin{eqnarray}
& Z_1 < Z_2 < Z_3 < \ldots < Z_m & 
\end{eqnarray}
In fact, Puget proves that we can eliminate
{\em all} value symmetry (and not just that due
to value interchangeability) with a linear
number of such ordering constraints. 
Unfortunately, this decomposition into 
ordering constraints
hinders propagation even for the tractable case of interchangeable
values (Theorem 5 in \cite{wcp07}). 
Indeed, even with \emph{just} two value symmetries,
mapping into variable symmetry can hinder propagation. 
This is supported by the experiments in \cite{wcp07}
where we see faster and more effective symmetry
breaking with the global \PRECEDENCE\ constraint. 
This global constraint thus appears to be a promising method to eliminate 
the symmetry due to interchangeable values. 

A generalization of the symmetry due to interchangeable
values is when values partition into sets, and values
within each set (but not between sets) are interchangeable.
The idea of value precedence can be generalized
to this case \cite{wecai2006}. 
The global $\GENPRECEDENCE$ 
constraint ensures that values
in each interchangeable set occur in order. 
More precisely, 
if the values are divided into $s$ equivalence classes,
and the $j$th equivalence class contains the
values $v_{j,1}$ to $v_{j,m_j}$ 
then $\GENPRECEDENCE$ ensures
$\min \{ i \ | \ X_i=v_{j,k} \vee i=n+1\} < 
 \min \{ i \ | \ X_i=v_{j,k+1} \vee i=n+2\}$ 
for all $1 \leq j \leq s$ and $1 \leq k < m_j$. 
Enforcing domain consistency
on $\GENPRECEDENCE$
is NP-hard in general but
fixed-parameter tractable
in $k = s$ \cite{wcp07,bhhkqwaaai2008}.

\section{Row and column symmetry}

Another common type of symmetry where we 
can exploit special properties of the symmetry group
is row and column symmetry 
\cite{ffhkmpwcp2002}.
Many problems can be modelled by a matrix model involving
a matrix of decision variables \cite{matrix2001,matrix2002,cp2003}.
Often the rows and columns
of such matrices are fully or partially interchangeable
\cite{ffhkmpwcp2002}.
For example, the Equidistant Frequency Permutation Array (EFPA) problem
is a challenging combinatorial problem in coding theory. The aim
is to find a set of $v$ code words, each of length $q\lambda$
such that each word contains $\lambda$ copies of the symbols
1 to $q$, and each pair of code words is at a Hamming distance
of $d$ apart. For example, for $v=4$, $\lambda=2$, $q=3$, $d=4$,
one solution is:
\begin{equation} \label{epfa}
\begin{array}{cccccc}
0 & 2 & 1 & 2 & 0 & 1 \\
0 & 2 & 2 & 1 & 1 & 0 \\
0 & 1 & 0 & 2 & 1 & 2 \\
0 & 0 & 1 & 1 & 2 & 2  
\end{array}
\end{equation}
This problem has applications in 
communications,
and is closely 
related to other combinatorial
problems like finding 
orthogonal Latin squares.
Huczynska {\it et al.} \cite{hmmncp09} consider a simple matrix 
model for this problem with a $v$ by $q\lambda$
array of variables, each with domains $1$ to $q$. 
This model has row and column symmetry since we can
permute the rows and the columns of 
any solution. Although breaking all row and
column symmetry is intractable in general, it 
is fixed-parameter tractable in the number of
columns (or rows). 

\begin{mytheorem}
With a $n$ by $m$ matrix,
checking lex leader constraints that break all row
and column symmetry
is NP-hard in general but
fixed-parameter tractable
in $k = m$.
%
\end{mytheorem}
\myproof
NP-hardness is proved by Theorem 3.2 in \cite{clgrkr96},
and fixed-parameter tractability by Theorem 1 in \cite{knwcp10}.
\myqed

Note that the above result only talks about
{\it checking} a constraint which breaks all row and
column symmetry. That is, we only consider the computational
cost of deciding if a complete assignment satisfies the constraint. 
Propagation of such a global
constraint is computationally more difficult. 

Just row or column symmetry on their own are
tractable to break. 
To eliminate all row symmetry we can
post lexicographical ordering constraints
on the rows. Similarly,
to eliminate all column symmetry we can 
post lexicographical ordering constraints
on the columns. 
When we have both row and column symmetry,
we can post 
a $\DLex$ constraint that 
lexicographically orders
both the rows and columns
\cite{ffhkmpwcp2002}.
This does not eliminate
all symmetry since
it may not break symmetries which permute 
both rows and columns. Nevertheless, it is 
more tractable to propagate and is often
highly effective in practice. 
Note that $\DLex$ can be derived
from a {\em strict} subset of the \LexLeader\ constraints.
Unfortunately propagating such 
a \DLex\ constraint completely is
already NP-hard. 

\begin{mytheorem}
With a $n$ by $m$ matrix,
enforcing domain consistency
on $\DLex$
is NP-hard in general. 
\end{mytheorem}
\myproof
See Threorem 3 in \cite{knwcp10}.
\myqed

There are two special 
cases of matrix models
where row and column symmetry is
more tractable to break. 
The first case is with an all-different matrix, a matrix model
in which every value is different.
If an all-different matrix has row and column symmetry
then the lex-leader method
ensures that the top left entry is the smallest value,
and the first row and column are ordered
\cite{ffhkmpwcp2002}. 
Domain consistency can be enforced on
such a global constraint in polynomial time
\cite{knwcp10}.
The second more tractable case is with a matrix model of a function.
In such a model, all entries are 0/1 and each row sum is 1. 
If a matrix model of a function has row and column symmetry
then the lex-leader method ensures
the rows and columns are lexicographically
ordered, the row sums are 1, 
and the sums of the columns are in decreasing
order \cite{ilya01,lex2001,ffhkmpwcp2002}.
Domain consistency can also be enforced on
such a global constraint in polynomial time
\cite{knwcp10}.

\section{Related work}

The study of computational complexity 
in constraint programming has tended to focus on the structure
of the constraint graph (e.g. especially 
measures like tree width \cite{freuder4,dechter7})
or on the semantics of the constraints (e.g. \cite{cooper2}).
However, these lines of research are mostly concerned with constraint
satisfaction problems as a whole, and do not say much about
individual (global) constraints.
For global constraints of bounded arity, asymptotic
analysis has been used to characterize 
the complexity of
propagation both in general and
for constraints with a particular semantics. For example,
the generic domain consistency algorithm of \cite{Bessiere-Regin97}
has an $O(d^n)$ time complexity on constraints
of arity $n$ and domains of size $d$,
whilst the domain consistency algorithm of  \cite{regin1}
for the $n$-ary \alldiff\
constraint has $O(n^{\frac{3}{2}}d)$
time complexity.
Bessiere {\it et al.}
showed that many global constraints like \nvalue\ are
also intractable to propagate \cite{bhhwaaai2004}. More recently,
Samer and Szeider have studied the parameterized
complexity of the \egcc\ constraint \cite{szeider2}.
Szeider has also studied the complexity of 
symmetry in a propositional resolution 
calculus \cite{stcs2005}. See Chapter 10 in \cite{bhmwhandbook}
for more about symmetry of propositional systems. 

\section{Conclusions}

We have argued that parameterized complexity is a useful
tool with which to study symmetry breaking.
In particular, we have shown that whilst
it is intractable to break all symmetry completely,
there are special types of symmetry like value
symmetry and row and column symmetry
which are more tractable to break. 
In these case, fixed-parameter
tractability comes from natural parameters like
the number of generators which tend to be small. 
In future, we hope that insights provided by such
analysis will inform the design
of new search methods. For example,
we might build a
propagator that propagates completely
when the parameter is small, but only partially
when it is large. 
In the longer term, we hope 
that other aspects of parameterized complexity
like kernels will find application in the domain
of symmetry breaking.

\bibliographystyle{splncs}





\end{document}